\documentclass[conference]{IEEEtran}
\ifCLASSOPTIONcompsoc
  \usepackage[nocompress]{cite}
\else
  \usepackage{cite}
\fi
\ifCLASSINFOpdf
   \usepackage[pdftex]{graphicx}
\else
\fi
\usepackage{amsmath}
\usepackage{algorithmic}
\ifCLASSOPTIONcompsoc
 \usepackage[caption=false,font=footnotesize,labelfont=sf,textfont=sf]{subfig}
\else
 \usepackage[caption=false,font=footnotesize]{subfig}
\fi
\usepackage{amssymb}
\usepackage{booktabs} 
\usepackage{microtype}
\usepackage{graphicx}
\usepackage{hyperref}
\usepackage{dirtytalk}
\usepackage{multirow,tabularx}

\usepackage{mathtools, nccmath}

\newcommand\blfootnote[1]{%
  \begingroup
  \renewcommand\thefootnote{}\footnote{#1}%
  \addtocounter{footnote}{-1}%
  \endgroup
}

\begin{document}
%
\title{Quantifying and Using System Uncertainty\\in UAV Navigation}



\author{
\IEEEauthorblockN{Fabio Arnez}
\IEEEauthorblockA{Universit\'e Paris-Saclay, CEA, List\\F-91120, Palaiseau, France\\
fabio.arnez@cea.fr}
\and
\IEEEauthorblockN{Ansgar Radermacher}
\IEEEauthorblockA{Universit\'e Paris-Saclay, CEA, List\\F-91120, Palaiseau, France\\
ansgar.radermacher@cea.fr}
\and
\IEEEauthorblockN{Huascar Espinoza}
\IEEEauthorblockA{KDT JU\\Brussels, Belgium\\
huascar.espinoza@kdt-ju.europa.eu}

}

\maketitle
\thispagestyle{plain}
\pagestyle{plain}

\begin{abstract}
As autonomous systems increasingly rely on Deep Neural Networks (DNN) to implement the navigation pipeline functions, uncertainty estimation methods have become paramount for estimating confidence in DNN predictions. Bayesian Deep Learning (BDL) offers a principled approach to model uncertainties in DNNs. However, DNN components from autonomous systems partially capture uncertainty, or more importantly, the uncertainty effect in downstream tasks is ignored. This paper provides a method to capture the overall system uncertainty in a UAV navigation task. In particular, we study the effect of the uncertainty from perception representations in downstream control predictions. Moreover, we leverage the uncertainty in the system's output to improve control decisions that positively impact the UAV's performance on its task.

\end{abstract}


%
\IEEEpeerreviewmaketitle


 



\blfootnote{Preprint version. \textbf{Accepted} at the \textit{ICRA 2022 Workshop on Releasing Robots into the Wild: Simulations, Benchmarks, and Deployment}.
Copyright 2022 by the author(s).}

\section{Introduction}
Navigation in complex environments still represents a big challenge for automated systems (AS). Particular instances of this problem are autonomous driving and autonomous aerial navigation in the context of Unmanned Aerial Vehicles (UAVs). In both cases, the navigation task is addressed by first acquiring rich and complex raw sensory information (e.g., from camera, radar, LiDAR, etc.), which is then processed to drive the agent towards its goal. Usually, this process is done in sequence, where tasks and specific software components are linked together in the so-called \textit{perception-planning-control} software pipeline \cite{siegwart2011introduction,mcallister2017concrete}. In the last decade, Deep Neural Networks (DNNs) have become a popular choice to implement navigation components. For this purpose, three paradigms exist to develop and train DNN-based components: Modular learning (isolated) \cite{grigorescu2020survey}, End-to-End (E2E) learning \cite{loquercio2018dronet,codevilla2019exploring,zeng2019end}, and a mixed approach \cite{bonatti2019learning,mueller2018driving}.

\begin{figure}[!t]
	\centering
	\includegraphics[width=0.7\linewidth]{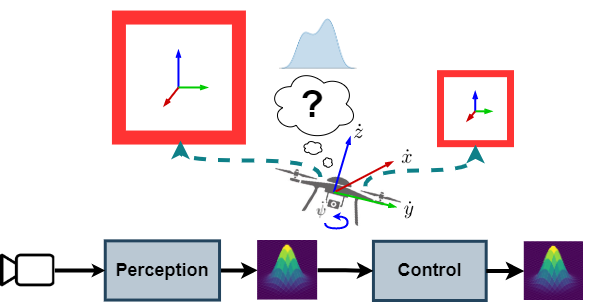}
	\caption{UAV Uncertainty-Aware Aerial Navigation Pipeline}
	\label{fig:simple-uncertainty-nav}
\end{figure}

Despite the remarkable progress in representation learning, DNNs should also represent the confidence in their predictions to deploy them in safety-critical systems. McAllister et al. \cite{mcallister2017concrete} proposed using Bayesian Deep Learning (BDL) to implement the components from navigation pipelines. Bayesian methods offer a principled framework to model and capture uncertainty. Nevertheless, if the Bayesian approach is followed, all the components in the system should use BDL methods to enable uncertainty propagation in the navigation pipeline. Hence, BDL components should admit uncertainty information as an input to account for the uncertainty from the outputs of preceding BDL components (See Fig.~\ref{fig:simple-uncertainty-nav} bottom).


In recent years, a large body of literature has employed uncertainty estimation methods in robotic tasks thanks to its potential to improve the safety of automated functions \cite{michelmore2018evaluating}, and the capacity to increase the task performance \cite{nozarian2020uncertainty, ohn2020learning}. However, uncertainty is captured partially in navigation pipelines that utilize DNNs. BDL methods are used mainly in perception task, and downstream components (e.g., planning and control) usually ignore the uncertainty from the preceding components or do not capture uncertainty in their predictions. Although some works propagate downstream perceptual uncertainty from intermediate representations \cite{arnez2021improving,ivanovic2021heterogeneous,casas2020implicit}, the overall system output does not take into account all the uncertainty sources from DNN components in the pipeline.



Quantifying uncertainty in a BDL-based system (i.e., a pipeline of BDL components) still remains a challenging task. Uncertainties from BDL components must be assembled in a principled way to provide a reliable measure of overall system uncertainty, based on which safe decisions can be made \cite{mcallister2017concrete,lavin2021technology,ruess2022safe} (See Fig.~\ref{fig:simple-uncertainty-nav} top). In this paper, we propose to capture and use the the overall system uncertainty in the navigation pipeline to improve the UAV performance on its task.

\section{System Task Formulation}

In this paper, we address the problem of autonomous aerial navigation. The goal of the UAV is to navigate through a set of gates with unknown locations disposed in a circular track. Following prior work from \cite{bonatti2019learning,arnez2021improving}, the navigation architecture consists of two neural networks: one for perception and the other for control, as shown in Fig.~\ref{fig:SystemArch}. To achieve goal, the navigation task is formulated as a sequential-decision making problem, where a control action is produced given an environment observation. In this regard, the simulation environment provides at time $t$ an observation comprised of an RGB image $\mathbf{x}$ acquired from a front-facing camera on the UAV. The perception component defines an encoder function  $q_{\phi}:\mathcal{X} \rightarrow \mathcal{Z}$ that maps the input image $\mathbf{x}$ to a rich low dimensional representation  $\mathbf{z} \in \mathbb{R}^{10}$. Next, a control policy $\pi_{w}: \mathcal{Z} \rightarrow \mathcal{Y}$ maps the compact representation $\mathbf{z}$, to control commands $\mathbf{y} = [\dot{x}, \dot{y}, \dot{z}, \dot{\psi}] \in \mathbb{R}^{4}$, corresponding to linear and yaw velocities in the UAV body frame.


In the perception component, a cross-modal variational autoencoder (CMVAE) \cite{spurr2018cross,bonatti2019learning} is used to learn a rich and robust compact representation. A CMVAE is a variant of traditional variational autoencoders (VAE) \cite{kingma2013auto} that learns a single latent representation for multiple data modalities. In this case, two data modalities are used: the RGB images and the pose of the gate relative to the UAV body-frame. The CMVAE encoder $q_{\phi}$ maps then an input image $\mathbf{x}$ to a noisy representation with mean $\mu_{\phi}(\mathbf{x})$ and variance $\sigma^{2}_{\phi}(\mathbf{x})$ in the latent space, from where latent vectors $z$ are sampled, $\mathbf{z} \sim \mathcal{N}(\mu_{\phi},\sigma^{2}_{\phi})$. The encoder $q_{\phi}$ is based on the Dronet architecture \cite{loquercio2018dronet}, and additional constraints on the latent space are imposed to promote the learning of robust disentangled representations. For the downstream control task (control policy $\pi$), a feed-forward network is used to operate on the latent vectors $\mathbf{z}$. For more information about the general architecture for aerial navigation, we refer the reader to \cite{bonatti2019learning,arnez2021improving}.

\section{Methodology}


\subsection{Uncertainty from Perception Representations}

Although the CMVAE encoder $q_{\phi}$ employs Bayesian inference to obtain latent vectors $\mathbf{z}$, CM-VAE does not capture epistemic uncertainty since the encoder lacks a distribution over parameters $\phi$. To capture uncertainty in the perception encoder we follow prior work from \cite{daxberger2019bayesian,jesson2020identifying} that attempts to capture epistemic uncertainty in VAEs. We adapt the CM-VAE to capture the posterior $q(\mathbf{z} \mid \mathbf{x}, \mathcal{D}_p)$ as shown in eq. \ref{eq:postEncoder}.

\begin{equation}
    q_{\Phi}(\mathbf{z} \mid \mathbf{x}, \mathcal{D}_{p}) = \int{q(\mathbf{z} \mid \mathbf{x}, \phi) p(\phi \mid \mathcal{D}_{p})d\phi}
	\label{eq:postEncoder}
\end{equation}

To approximate eq. \ref{eq:postEncoder} we take a set ${\Phi = \{\phi_{m}\}^{M}_{m}}$ of encoder parameters samples $\phi_{m} \sim p(\phi \mid \mathcal{D}_{p})$, to obtain a set of latent samples $\{\mathbf{z}_{m}\}^{M}_{m=1} \sim q_{\Phi}(\mathbf{z} \mid \mathbf{x}, \mathcal{D}_{p})$ at the output of the encoder. In practice, we modify CMVAE by adding a dropout layer in the encoder. Then, we use Monte Carlo Dropout (MCD) \cite{gal2016dropout} to approximate the posterior on the encoder weights $p(\phi \mid \mathcal{D}_{p})$. Finally, for a given input image $\mathbf{x}$ we perform $M$ stochastic forward passes (with dropout \say{turned on}) to compute a set of $M$ latent vector samples $\mathbf{z}$ at runtime.

\begin{figure}[!t]
	\centering
	\includegraphics[width=0.99\linewidth]{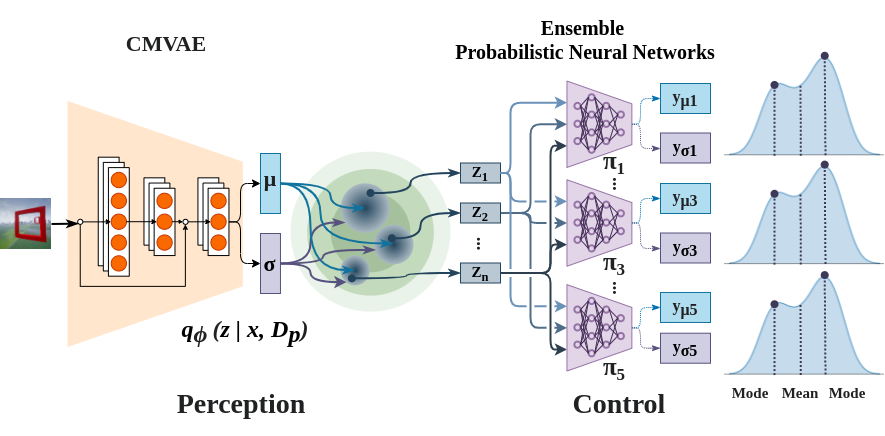}
	\caption{System architecture for aerial navigation}
	\label{fig:SystemArch}
\end{figure}

\subsection{Input Uncertainty for Control}

In BDL, downstream uncertainty propagation assumes that a neural network component is able to handle or admit uncertainty at the input. In our case, this implies that the neural network for control is able to handle the uncertainty coming from the perception component.
When BNNs are used with LV (BNN+LV), both types of uncertainty can be captured \cite{depeweg2017learning, depeweg2018decomposition,henaff2018model,arnez2021improving}.
To capture the overall system uncertainty at the output of the controller, we compute the posterior predictive distribution for target variable $\mathbf{y^*}$ associated with a new input image $\mathbf{x^*}$, as shown in eq. \ref{eq:postPredDist}.


\begin{multline}
    p(\mathbf{y^*} \mid \mathbf{x^*}, \mathcal{D}_{c}, \mathcal{D}_{p}) = \\
	        \iint{\pi(\mathbf{y} \mid \mathbf{z}, \mathbf{w}) p(\mathbf{w} \mid \mathcal{D}_{c})q_{\Phi}(\mathbf{z} \mid \mathbf{x^{*}}, \mathcal{D}_{p}) d\mathbf{w} d\mathbf{z}}
	\label{eq:postPredDist}
\end{multline}

The above integrals are intractable, and we rely on approximations to obtain an estimation of the predictive distribution. The posterior $p(\mathbf{w} \mid \mathcal{D}_{c})$ is difficult to evaluate, thus we can approximate the inner integral using an ensemble of neural networks \cite{gustafsson2019evaluating}. In practice, we train an ensemble of $N$ probabilistic control policies $\{ \mathbf{\pi}_{n}(\mathbf{y} \mid \mathbf{z}, \mathbf{w}_{n})\}_{n=1}^{N}$, with weights $\{\mathbf{w}_{n}\}^{N}_{n=1} \sim p(\mathbf{w | \mathcal{D}})$, and where each control policy $\pi_{n}$ in the ensemble predicts the mean $\mu$ and variance $\sigma^{2}$ for each velocity command, i.e.,
$\mathbf{y}_{\mu} = [\mu_{\dot{x}}, \mu_{\dot{y}}, \mu_{\dot{z}}, \mu_{\dot{\psi}}]$ and
$\mathbf{y}_{\sigma^{2}} = [\sigma^{2}_{\dot{x}}, \sigma^{2}_{\dot{y}}, \sigma^{2}_{\dot{z}}, \sigma^{2}_{\dot{\psi}}]$. For training each probabilistic control policy we use imitation learning and the heteroscedastic loss function, as suggested by \cite{kendall2017uncertainties, lakshminarayanan2017simple}.

The outer integral is approximated by taking a set of samples from the perception component latent space. In \cite{arnez2021improving} latent samples are drawn using the encoder mean and variance~$\mathbf{z} \sim \mathcal{N}(\mu_{\phi},\sigma^{2}_{\phi})$. For the sake of simplicity, we directly use the samples obtained in the perception component~$\{\mathbf{z}_{m}\}^{M}_{m} \sim q_{\Phi}(\mathbf{z} \mid \mathbf{x}, \mathcal{D}_{p})$ to take into account the epistemic uncertainty from the previous stage. Finally, the predictions that we get from passing each latent vector $\mathbf{z}$ through each ensemble member are used to estimate the posterior predictive distribution in eq. \ref{eq:postPredDist}. From the control policy perspective, using multiple latent samples $\mathbf{z}$ can be seen as taking a better \say{picture} of the latent space (perception representation) to gather more information about the environment. Interestingly, we can also make a connection between our sampling approach and MEMO \cite{zhang2021memo}, a method that performs augmentations on test dataset points to improve prediction robustness. Our  method is illustrated in Fig. \ref{fig:SystemArch}.

\subsection{Decision-Making Under Uncertainty}

For control, assigning the same weight to each ensemble member prediction to compute the ensemble mixture predictive mean and variance \cite{lakshminarayanan2017simple, gustafsson2019evaluating} can result in sub-optimal solutions when facing multimodal predictions that could arise from ambiguous inputs. For example, the predictions from one control policy network could attempt to move the UAV to the left, while the predictions from another control policy could try to move the UAV to the right. This can result in the UAV moving straight and hence can lead to a tragic result.


To overcome this problem, we take inspiration from techniques on active learning for Bayesian Deep Learning \cite{gal2017deep,kirsch2019batchbald}, that use mutual information. However, in our approach, given the samples from the latent space, we propose to choose the control predictions (the density) from the ensemble member that minimizes the mutual information, as presented in eq. \ref{eq:milb_min_sel}.

\begin{equation}
    {\pi}^{*} = \underset{\pi}{\arg\min} \big( \mathbb{I}(y ; z, w) \big), \forall y \in \mathbf{y} = [\dot{x}, \dot{y}, \dot{z}, \dot{\psi}]
    \label{eq:milb_min_sel}
\end{equation}

In our navigation architecture, the mutual information $\mathbb{I}$ is formulated as follows,

\begin{equation}
    \mathbb{I}(y ; z, w) = \int_{z}{\mathbf{KL} \big( \pi(y \mid z, w) \parallel \pi(y) \big) p(z) dz}
\end{equation}
To estimate the mutual information from the previous integral, we use the variational lower bound approximation from \cite{poole2019variational} taking the latent samples $\mathbf{z}$ and each ensemble member predictions ($\hat{\mu}$ and $\hat{\sigma}$), as presented in eq. \ref{eq:milb}:
\begin{multline}
    \mathbb{I}(y ; z, w) \geqslant \\
	        \mathbb{E}_{z_{1:M}} \Big[ \sum_{i=1}^{M}{ \mathbf{KL} \big( \pi(y \mid z, w) \parallel m(y ; z_{1:M}, w_{1:N}) \big)} \Big]
	\label{eq:milb}
\end{multline}
Once an ensemble member is chosen, we can use the control policy predicted densities by taking the mean or the modes, as presented in Fig.\ref{fig:SystemArch}.








\section{Experiments}
For our experiments, we study the impact of the uncertainty from perception representations in a downstream control component. We seek to answer the following research questions. \textbf{RQ1.}~How does uncertainty from perception representations affect the downstream control predictions? \textbf{RQ2.}~Can we improve the UAV performance using perception and control uncertainties?

\begin{figure}[t!]
\centering
    \subfloat[\centering Circular track view]{\includegraphics[width=0.45\linewidth]{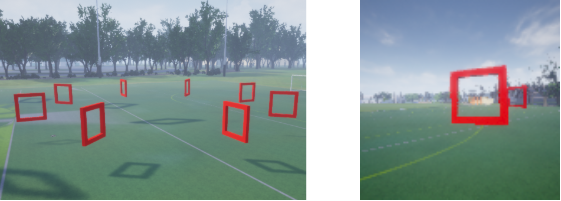}}
    \hfill
    \subfloat[\centering Noisy circular track view]{\includegraphics[width=0.45\linewidth]{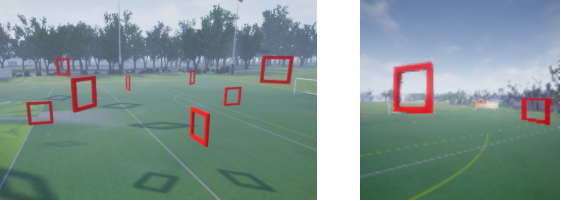}}
\caption{Aerial navigation tracks: birds-eye view (left), and view from UAV perspective (right)}
\label{fig:TrackViews}
\end{figure}

\subsection{Experimental setup}

\subsubsection{Navigation Model Baselines}
All the navigation architectures are based on \cite{bonatti2019learning} and are implemented using PyTorch. Table \ref{table:exp-nav-models} shows the uncertainty-aware navigation architectures used in our experiments, and the type of perception component, the number latent variable samples (LVS), the type of control policy, and the number of control prediction samples (CPS) at the output of the system. For instance, $\mathcal{M}_0$ represents our Bayesian navigation pipeline. $\mathcal{M}_0$ perception component captures epistemic uncertainty using MCD with 32 forward passes for each input to get 32 latent variable predictions. For the sake of simplicity, perception predictions are directly used as latent variable samples in downstream control. The control component uses a ensemble of 5 probabilistic control policies obtaining 160 control prediction samples.

\begin{table}[!t]
\caption{Uncertainty-aware Navigation models in the experiments}
\label{table:exp-nav-models}
\centering
\begin{tabular}{@{}cllll@{}}
\toprule
\textbf{Model} & \textbf{Perception $(q_{\phi})$} & \textbf{LVS} & \textbf{Control Policy $(\pi)$} & \textbf{CPS} \\ \midrule
$\mathcal{M}_0$ & MCD-CMVAE & 32 & Ensemble (5) Prob. & 160 \\
$\mathcal{M}_1$ & CMVAE     & 32 & Ensemble (5) Prob. & 160 \\
$\mathcal{M}_2$ & CMVAE     & 1  & Ensemble (5) Prob. & 5   \\
$\mathcal{M}_3$ & CMVAE     & 32 & Deterministic          & 32  \\
$\mathcal{M}_4$ & CMVAE     & 1  & Prob.          & 1   \\ \bottomrule
\end{tabular}
\end{table}

\begin{figure*}[t!]
\centering
    \subfloat[\centering Navigation model input, Perception prediction densities, and Control policy ensemble prediction densities ($\hat{\mu}, \hat{\sigma}$)]
        {
        \includegraphics[width=0.85\linewidth]{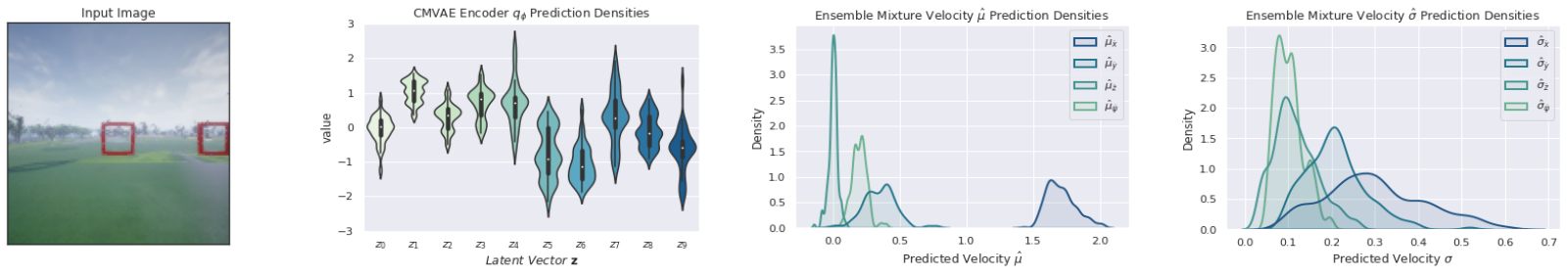}
        }

    \subfloat[\centering Control prediction densities  ($\hat{\mu}_{\dot{y}},\hat{\sigma}_{\dot{y}}$) and ($\hat{\mu}_{\dot{\psi}}, \hat{\sigma}_{\dot{\psi}}$) for each ensemble member]
        {
        \includegraphics[width=0.88\linewidth]{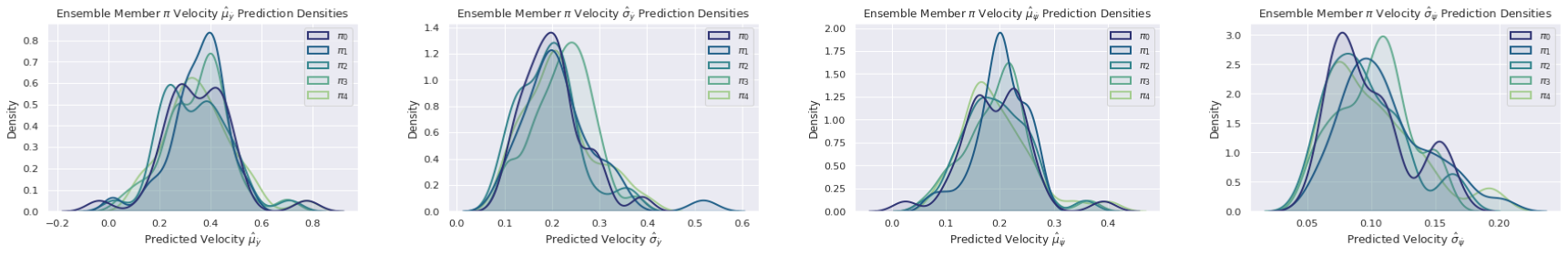}
        }
\caption{Bayesian Navigation model: Component prediction densities}
\label{fig:NavModelPreds}
\end{figure*}


\subsubsection{Datasets}
We use two independent datasets for each component in the navigation pipeline. The CMVAE uses a dataset of 300k images with gate pose labels. The control component uses a dataset of 17k images with drone velocity labels. The perception dataset split divided into 80\% for training, and the remaining 20\% for validation and testing. The control dataset uses a split of 90\% for training and the remaining for validation and testing. In both cases the image size is 64x64 pixels.

\subsubsection{Evaluation Procedure}

First, we observe $\mathcal{M}_0$ predictions with an out-of-distribution input sample that introduces ambiguity to the model (double-gate). Then, we evaluate the navigation model using AirSim. We use a circular track with eight equally spaced gates positioned initially in a radius of 8 m and constant height. To assess the system performance to perturbations in the environment, we generate new tracks adding random noise to each gate radius and height.
In the context of the AirSim \cite{shah2018airsim} simulation environment, a track is entirely defined by a set of gates, their poses in three-dimensional space, and the agent navigation direction. For perception-based navigation, the complexity of a track resides in the \say{\textit{gate-visibility}} difficulty \cite{madaan2020airsim,song2021autonomous}, i.e., how well the camera Field-of-View (FoV) captures the gate. A natural way to increase track complexity is by adding a random displacement to the position of each gate. A track without random displacement in the gates has a circular fashion. Gate position randomness alters the shape of the track, affecting the gate visibility, i.e., gates are: not visible, partially visible or multiple gates can be captured in the UAV FoV as presented in Fig.~\ref{fig:TrackViews}.
We measure the system performance by taking the average number of gates passed in six different tracks. The UAV mission considers a maximum of 32 gates which is equivalent to 4 laps (8 gates / lap), and two levels of noise for gate offset:  Gate Radius Noise (GRN) and Gate Height Noise (GHN). Each navigation model has two trials on each track. In addition, we consider two control decision-making strategies (CDMS), to use control predictions. The first strategy uses the deep ensemble mean. The second control strategy uses mutual information lower bound to select the ensemble member predicted density from where we choose the lowest velocity modes for $\dot{y}, \dot{z},\dot{\psi}$, while for $\dot{x}$  we choose a conservative strategy by selecting the lowest predicted velocity.

\begin{table}[!t]
\caption{Uncertainty-Aware Navigation Models Performance:\\Avg. number of gates passed}
\label{table:nav-models-exp-results}
\centering
\begin{tabular}{@{}cccc@{}}
\toprule
\multirow{2}{*}{\textbf{Model}} &
  \multirow{2}{*}{\textbf{CDMS}} &
  \multicolumn{2}{c}{\textbf{Passed Gates with Noise Level}} \\ \cmidrule(l){3-4} 
 &
   &
  \multicolumn{1}{l}{\textbf{\begin{tabular}[c]{@{}l@{}}$GRN=1.0$\\ $GHN=2.0$\end{tabular}}} &
  \multicolumn{1}{l}{\textbf{\begin{tabular}[c]{@{}l@{}}$GRN=1.5$\\ $GHN=3.0$\end{tabular}}} \\ \midrule
$\mathcal{M}_0$ & MI-mode & 28.88 & 23.05 \\
$\mathcal{M}_0$ & DE-mean & 19.77 & 9.22  \\
$\mathcal{M}_1$ & DE-mean & 17.67 & 6.0   \\
$\mathcal{M}_2$ & DE-mean & 17.33 & 4.0   \\
$\mathcal{M}_3$ & DE-mean & 8.33  & 5.0   \\
$\mathcal{M}_4$ & DE-mean & 15.16 & 4.38  \\ \bottomrule
\end{tabular}
\end{table}


\subsection{Results}

Fig. \ref{fig:NavModelPreds}(a) shows $\mathcal{M}_0$ predictions at the output of the perception ($\mathbf{z}$) and control ($\hat{\mu}$, $\hat{\sigma}$) components. Predictions are made using an input sample image with two gates to see if the model is able to capture the ambiguity from the sample. Interestingly, the control densities, using all the CPS from the ensemble, show that $\mathcal{M}_0$ is able to represent the ambiguity in the input. The predicted velocity densities shows complex multimodal distributions (two peaks) for $\hat{\mu}_{\dot{y}}$, $\hat{\mu}_{\dot{\psi}}$ and $\hat{\sigma}_{\dot{\psi}}$. the multimodal predictions are observed more in detail in Fig. \ref{fig:NavModelPreds}(b), in the prediciton densities for each ensemble member for $\dot{y}$ and $\dot{\psi}$ velocity commands.


Finally, Table \ref{table:nav-models-exp-results} presents the navigation performance results. In general, learning to predict uncertainty in the control components can boost the performance significantly. In the case of $\mathcal{M}_0$, the strategy to select control predictions clearly impacts the navigation performance. In general, using the control ensemble mean can lead to similar results. However, exploiting the ensemble members predicted distributions (as seen \ref{fig:NavModelPreds}) can boost the model performance.



\section{Conclusion}
We presented a method to capture and propagate uncertainty along a navigation pipeline implemented with Bayesian deep learning components for UAV navigation. We analyzed the effect of uncertainty propagation regarding system component predictions and performance. Our experiments show that our approach for capturing and propagating uncertainty along the system can provide valuable predictions and uncertainty estimates to build dependable systems. However, a proper use of component predictions and uncertainty estimates are needed to positively impact the system performance. In future work, we aim to explore sampling-free methods for uncertainty estimation \cite{charpentier2021natural} to reduce the computational budget and memory footprint in our approach.

\ifCLASSOPTIONcompsoc
  \section*{Acknowledgments}
\else
  \section*{Acknowledgment}
\fi
This work has received funding from the COMP4DRONES project, under Joint Undertaking (JU) grant agreement N\textdegree  826610. The JU receives support from the European Union's Horizon 2020 research and innovation programme and from Spain, Austria, Belgium, Czech Republic, France, Italy, Latvia, Netherlands.



%



\bibliographystyle{IEEEtran}
\bibliography{my_bib}
\end{document}